%% file: main.tex
\documentclass[letterpaper, 10 pt, conference]{ieeeconf}  

\IEEEoverridecommandlockouts                              

\input{misc/import}
\input{misc/settings}
\input{misc/math_abbrev}


\input{misc/thm_env}

\title{\LARGE \bf
On the Conic Complementarity of Planar Contacts
}

\author{Yann de Mont-Marin$^{\text{a},*}$, Louis Montaut$^{\text{a}}$, Jean Ponce$^{\text{a,b}}$, Martial Hebert$^{\text{c}}$, Justin Carpentier$^{\text{a}}$%
\thanks{$^{\text{a}}$Inria and Département d’Informatique de l’École Normale Supérieure, PSL Research University in Paris, 75013 Paris, France.}%
\thanks{$^{\text{b}}$Courant Institue and Center for Data Science, New York University, United States}%
\thanks{$^{\text{c}}$School of Computer Science, Carnegie Mellon University, United States}%
\thanks{*Corresponding author:
        {\tt\small yann.de-mont-marin@inria.fr}}%
}

\begin{document}

\maketitle
\thispagestyle{empty}
\pagestyle{empty}

\begin{abstract}
\input{tex/abstract}
\end{abstract}

\input{tex/1_introduction}
\input{tex/related-work}
\input{tex/2_contact_cop}
\input{tex/3_cone_complementarity}
\input{tex/5_conclusion}

\section*{Acknowledgments}
This work has received support from the French government, managed by the National Research Agency, under the France 2030 program with the references Organic Robotics Program (PEPR O2R), the PIQ program under the management of Agence de Programme du Numérique and “PR[AI]RIE-PSAI” (ANR-23-IACL-0008).
The European Union also supported this work through the ARTIFACT project (GA no.101165695) and the AGIMUS project (GA no.101070165).
Views and opinions expressed are those of the author(s) only and do not necessarily reflect those of the funding agencies.

\bibliographystyle{IEEEtranS}
\bibliography{bib}
\newpage
\section{Appendix}
\input{tex/proofs}
\end{document}

%% file: misc/import.tex
\usepackage{tabularx}
\usepackage{iftex}
\usepackage{times}
\usepackage{multicol}
\usepackage{url}
\usepackage{graphics} 
\usepackage{epsfig} 
\usepackage{xpatch}
\usepackage{xstring}
\usepackage{xspace}
\usepackage[table]{xcolor}
\usepackage[vlined,lined,ruled,linesnumbered,procnumbered,resetcount]{algorithm2e}
\usepackage{bm}
\usepackage{mathtools}
\usepackage{amsmath}
\usepackage{fouridx}
\usepackage{amsfonts}
\usepackage{amssymb}  
\usepackage{mathptmx} 
\usepackage[mathcal]{euscript}
\usepackage{mdframed}
\usepackage{booktabs}
\usepackage{esint}
\usepackage{bookmark}
\usepackage{cancel}
\usepackage{graphicx}
\usepackage{tikz}
\usepackage{tikz-cd}
\usetikzlibrary{backgrounds}
\usepackage{rotating}
\usepackage{stackengine}
\usepackage{tabularray}
\usepackage{wrapfig}
\usepackage{epigraph}
\usepackage{yhmath}
\usepackage{setspace}
\usepackage[most]{tcolorbox}
\usepackage{etoolbox} 
\usepackage{musicography}
\usepackage{textcomp}
\usepackage{lettrine}

%% file: misc/math_abbrev.tex
\newcommand{\RR}{\ensuremath{\mathbb{R}}\xspace}



%% file: misc/thm_env.tex
\newtheorem{lemma}{Lemma}
\newtheorem{proposition}{Proposition}



%% file: tex/abstract.tex
We present a unifying theoretical result that connects two foundational principles in robotics: the Signorini law for point contacts, which underpins many simulation methods for preventing object interpenetration, and the center of pressure (also known as the zero-moment point), a key concept used in, for instance, optimization-based locomotion control.
Our contribution is the planar Signorini condition, a conic complementarity formulation that models general planar contacts between rigid bodies.
We prove that this formulation is equivalent to enforcing the punctual Signorini law across an entire contact surface, thereby bridging the gap between discrete and continuous contact models. 
A geometric interpretation reveals that the framework naturally captures three physical regimes~—sticking, separating, and tilting—~within a unified complementarity structure. 
This leads to a principled extension of the classical center of pressure, which we refer to as the \textbf{extended center of pressure}. 
By establishing this connection, our work provides a mathematically consistent and computationally tractable foundation for handling planar contacts, with implications for both the accurate simulation of contact dynamics and the design of advanced control and optimization algorithms in locomotion and manipulation.

%% file: tex/1_introduction.tex
\section{Introduction}
\label{sec:intro}

The Signorini law for punctual contact is fundamental to contact modeling in robotics, mechanics, and computer graphics.
It formalizes rigid, frictionless, point contact as a nonpenetration condition expressed via complementarity between the gap and the normal contact force~\cite{acary2011}.

For a given contact point between two objects in contact, this law states that if a force acts on the contact point, it should be repulsive, and the contact velocity can only separate the objects in contact; however, the two cannot occur simultaneously.
Mathematically, it reads as a conic complementarity problem of the form:
\begin{align}
    \label{eq:signorini_condition_as_ccp}
    0\leq \rho_N \perp \nu_N \geq 0,
\end{align}
where $\rho_N$ is the normal contact force between the bodies, and $\nu_N$ is the normal relative velocity between the bodies.
These two quantities should be nonnegative (hence the $\geq 0$), and cannot be nonzero at the same time, encoded by the complementary operator $\perp$ that reads $\rho_N\nu_N = 0$.

This conic complementarity formulation is crucial for formulating contact dynamics as a conic optimization problem. 
Such optimization problems can then be easily exploited by dedicated optimization solvers, as done in many modern simulation software packages~\cite{todorov2012, lelidec2023, simplecontacts2024}. 
However, a conic complementarity formulation does not exist for a general planar contact patch composed of many, if not an infinite, number of contact points. 
The establishment of a conic formulation is the topic of this paper.\\

\begin{figure}[t!]
  \begin{center}
    \includegraphics[width=.6\linewidth, trim={0pt 0pt 0pt 0pt}, clip]{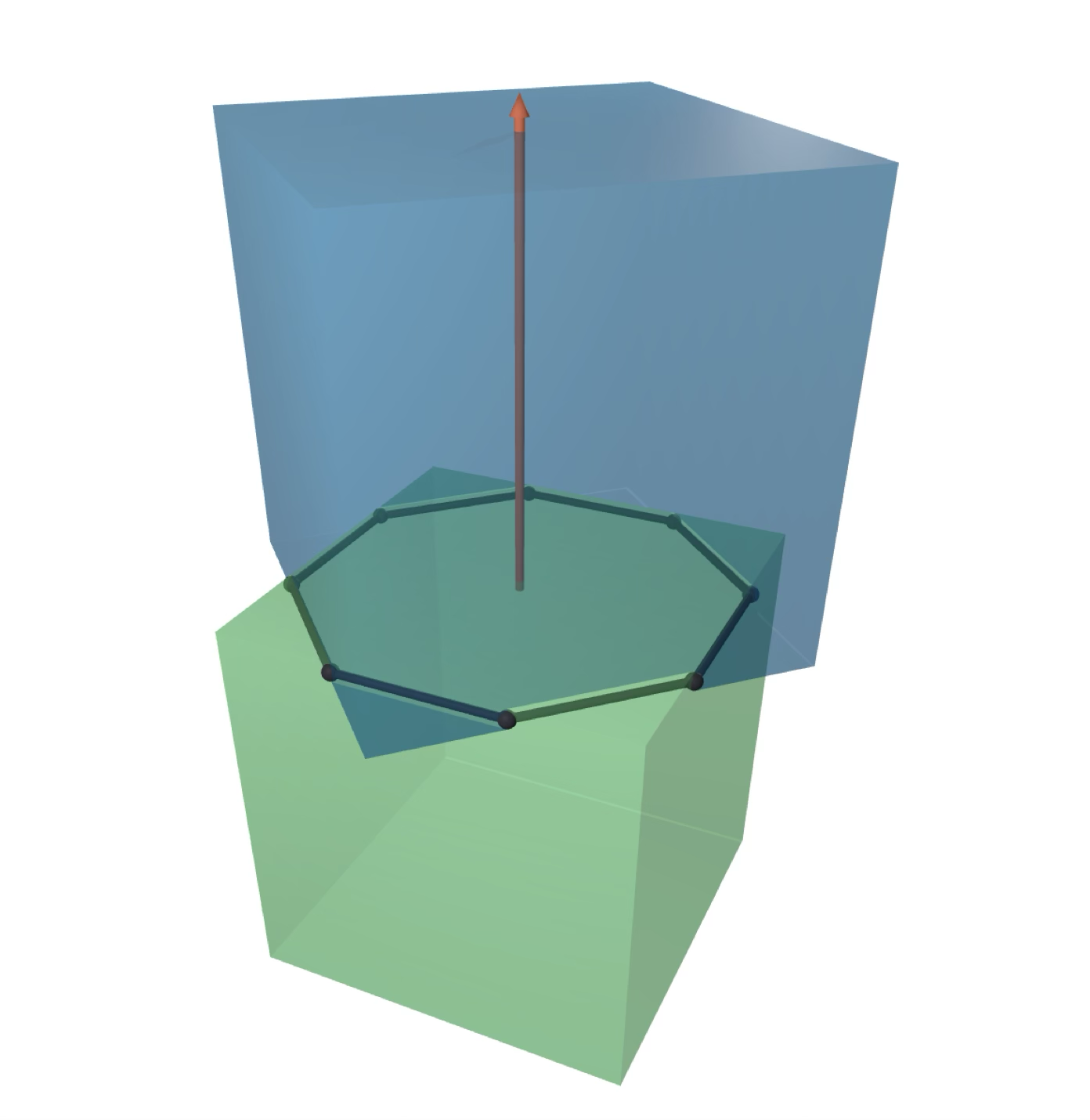}
    \vspace{-.6em}
    \caption[Illustration of a planar contact patch between two cubes.]{\textbf{Illustration of a planar contact patch between two cubes.}
    The contact surface is the intersection between the two faces, and the normal contact force distribution over this contact surface prevents the bodies from interpenetrating.
 }
    \label{fig:example-contact-patch}
  \end{center}
  \vspace{-1em}
\end{figure}

 When considering planar contact patches, a well-known result in control theory states that, due to the normal contact force distribution being positive, integrated over the contact surface, the so-called center of pressure must lie on the convex hull of the contact surface.
This result has notably been leveraged as a constraint to enforce when controlling a bipedal robot walking on a planar surface~\cite{kajita2003biped}.
Yet, this result only accounts for the positive pressure field associated with the Signorini condition, considering the presence of multiple points on the contact surface.\\

In this paper, we derive a complementarity formulation of any planar contact between rigid bodies.
We notably show that the Signorini condition applied at each point of a contact surface patch $P$ is equivalent to a conic complementarity that we call the \textit{planar Signorini condition}. 
Mathematically, this planar Signorini condition can be written as the conic complementarity constraint independent of the coordinate frame:
\begin{align}
    \label{eq:planar_signorini_condition}
    K_P \ni [\bm{m}_T,f_N] \perp [\bm{\omega}_T,v_N] \in K_P^*,
\end{align}
where $\bm{m}_T$ and $f_N$ are the moment and the resultant of the normal contact force distribution, whereas dually $\bm{\omega}_T$ and $v_N$ are the components of the twist responsible for the normal velocity field\footnote{In this paper, we use bold symbols to represent vector quantities, meaning that $x$ is a scalar quantity while $\bm{x} $ represents a vector.}. $K_P \subset \mathbb{R}^3$ is a cone, and $K_P^*$ is its dual cone.
In particular, we show later in the paper that the cone $K_P$ only depends on the shape of the planar contact patch $P$. 
Unlike in Eq.\eqref{eq:signorini_condition_as_ccp}, here the complementary operator $\perp$ applies to vectors, meaning that for two vectors $\bm{x}$ and $\bm y$ of $\mathbb{R}^3$, the complementarity reads $\langle\bm{x},\bm{y}\rangle = 0$. \\

Overall, this paper contributes to bridging the gap between the punctual Signorini condition, fundamental in simulation, and the center of pressure constraint, fundamental in biped control.
We establish that the planar Signorini condition extends the result that the center of pressure belongs to the patch. We demonstrate that it fully encompasses the three physical regimes of planar contact — sticking, separating, and tilting — within a single and unique condition.
We make these three main contributions:
\begin{itemize}
    \item We present a derivation of the conic complementarity associated with planar contact, that we name \textit{the planar Signorini condition}.
    \item We prove the equivalence of the punctual Signorini condition for every point of the contact surface and the planar Signorini condition.
    \item We provide a geometric interpretation of the planar Signorini condition, extending the classic interpretation to the center of pressure. We propose to name this extension the \textit{extended center of pressure}.\\
\end{itemize}

The paper is organized as follows.
Section~\ref{sec:related-work} provides a review of the related work.
In Section~\ref{sec:contact-cop}, we recall the punctual Signorini law and define the center of pressure, its relation to the moment of the normal contact force distribution, and the usual result that it lies on the surface of contact. In Section~\ref{sec:construction}, we formulate this result as a cone constraint, and we investigate the constraint on the normal velocity field to develop the dual constraint and obtain the cone complementarity formulation.
This section focuses on the construction, and the geometric interpretation and the formal proof are given in the Appendix.
Section~\ref{sec:conclusion} concludes the paper and highlights some potential applications of the planar Signorini condition for simulation and control.

%% file: tex/related-work.tex
\section{Related Work}
\label{sec:related-work}

\noindent\textbf{Contact simulation.}
Since the 1970s, extensive research has focused on simulating the dynamics of systems subject to both generalized forces and contacts, with or without friction~\cite{Panagiotopoulos1975, Katona1983, delpiero195, Moreau1988, Anitescu2010, Haslinger2012}.
%
Typically, modern simulation softwares used in robotics (MuJoCo, Drake, Bullet, PhysX, Newton, Genesis, etc.) employ a discrete frictional contact problem, where the system has contacts at a finite number of contact points. 
For each contact point, the associated contact force adheres to the Signorini condition, encoding that the rigid bodies cannot penetrate further, the contact force is repulsive, and these two conditions cannot occur simultaneously, as well as a friction model, such as the Coulomb model~\cite{acary2018,lelidec2023}.
Such a problem, together with the dynamic equations, can be cast as a complementarity problem and, in turn, as an optimization problem, allowing the use of efficient solvers~\cite{todorov2012, lelidec2023, simplecontacts2024, carpentier2021} to compute the contact forces and velocities that fulfill the contact modeling hypothesis.

In this setup, planar contacts are modeled using a finite number of points (typically 4) defining the boundary of the contact surface, creating an optimization problem of very high dimension and poor conditioning~\cite{cadoux2009}, which can significantly slow down the simulation if not adequately solved, using, for instance, proximal methods~\cite{simplecontacts2024}.

Some ad hoc algorithms~\cite{bouchard2015, xie2016, selig2011} have been dedicated to treating patch contact for robotic systems. Still, none of them model the proper complementarity of planar contact, which enables treating them as efficiently as point contact using the well-studied optimization arsenal~\cite{Boyd2004,boydProximalAlgorithms,wright1999numerical}.
The conic complementarity presented in this paper addresses this fundamental gap.

The work presented in~\cite{acary2011} remains an exception, casting punctual contact with rolling constraints as a nonlinear complementarity problem. 
However, it only partially models patch contact, as the rolling constraint implicitly arises from frictional constraints on a small patch of the surface. 
Still, the Signorini condition is only enforced at a single point, not over the entire patch.\\

\noindent\textbf{Legged robot control.}
The constraint that the center of pressure (CoP) must remain within the convex hull of the contact surface has been fundamental to bipedal robot stability analysis and control since the early development of humanoid robots.
Sardain et al. demonstrated this principle in their anthropomorphic biped robot design, establishing that maintaining the CoP within the support polygon is essential for static equilibrium and dynamic stability~\cite{Sardain98}.
This constraint was further validated in practical implementations such as the BIP2000 robot, where Espiau and Sardain showed that violating this geometric boundary leads to inevitable tipping and loss of balance~\cite{Espiau00}.

The seminal work of Kajita on the locomotion of bipedal systems~\cite{kajita2003biped} extensively relies on leveraging constraints on the zero-moment point~\cite{vukobratovic2004zero}, this point coinciding with the center of pressure in static conditions.
Azevedo et al. extended this framework by incorporating CoP trajectory planning within the support polygon as a critical component of their gait design methodology, demonstrating how the constraint directly influences step timing and foot placement strategies~\cite{Azevedo04}.

Wieber provides a rigorous formalization of this constraint within the context of walking system stability, proving that the CoP-support polygon relationship serves as a necessary condition for maintaining dynamic equilibrium during locomotion~\cite{wieber2002}.
This geometric constraint remains a cornerstone of modern bipedal control algorithms, with contemporary approaches building upon these foundational works to develop more sophisticated stability criteria and control strategies.

\begin{figure}[t!]
  \vspace{1em}
  \centerline{%
    \includegraphics[width=.9\linewidth, trim={0pt 0pt 30pt 0pt}, clip]{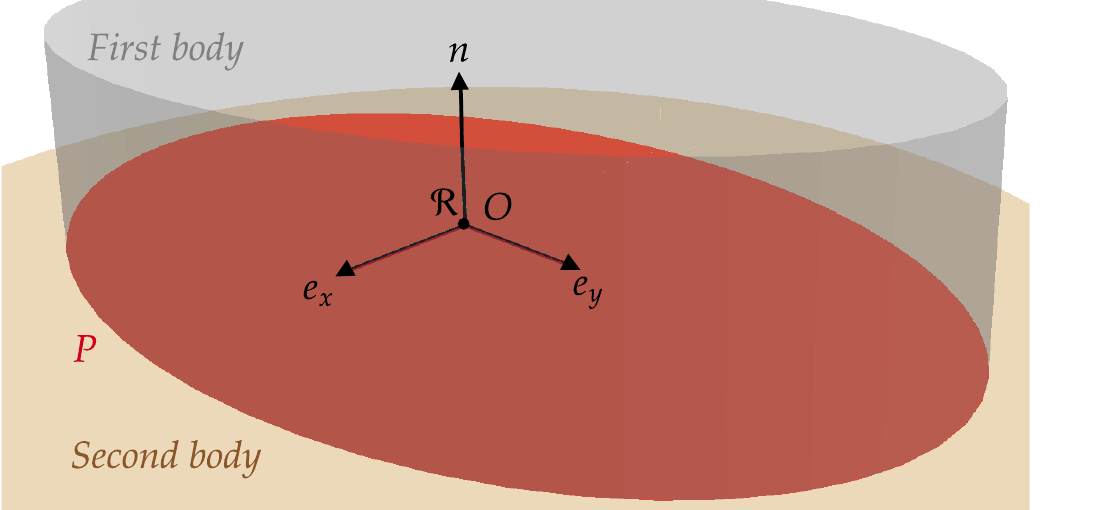}
 }
  \caption[Planar contact between two bodies.]{\textbf{Planar contact between two bodies.}
  The planar contact patch of coordinate $P$ in $\RR^2$ is supported by the plane $(O,e_x,e_y)$.
  The third axis of the frame $\mathcal{R}$ is the normal of the contact.
  For this figure and the following ones, we use an elliptical patch for illustration purposes.
  However, the results hold for any planar contact patch, even nonconvex ones.
  }
  \label{fig:setup-planar}
\end{figure}

%% file: tex/2_contact_cop.tex
\section{Background}
\label{sec:contact-cop}

In this section, we review the foundational concepts that underpin our formulation of planar contact: the notion of a contact patch, the Signorini condition for point contacts, and the center of pressure in the context of distributed contact forces. 
The contact patch, representing the physical region of interaction, serves as the bridge between discrete and continuous models.
The Signorini condition provides a complementarity framework for modeling point contacts between rigid bodies, forming the basis for many contact simulation methods. The center of pressure, commonly used in locomotion and balance control, summarizes the effect of distributed normal forces over a contact area.
Together, these concepts provide the theoretical context for the extended formulation developed in this work.

\subsection{Contact patch}
\label{subsec:patch-contact}

Consider a planar contact occurring between two bodies. The construction is independent of the coordinate frame, but for clarity of the presentation, we choose a frame $\mathcal{R}$ such that the origin and the first two components define the contact plane and the third component corresponds to the normal to the contact as depicted in Fig.~\ref{fig:setup-planar}.
We denote by $P$, a subset of $\RR^2$, the frame coordinates of a regular compact contact patch of the contact plane.

The instantaneous motion between the two rigid bodies is given by a twist $\sigma \in \mathfrak{se}(3)$.
In practice, the relative twist between two bodies of a poly-articulated system is given by the forward kinematics \cite{featherstone2008}.
Using the frame, we decompose $\sigma = [\bm\omega, \bm v] = [\bm\omega_T, \omega_N, \bm{v}_T, v_N]$ to distinguish the tangential component (in $\RR^2$) and normal component (in $\RR$) of angular and linear part of the twist.
The twist induces a velocity vector field on the contact plane $\nu:\RR^2\rightarrow\RR^3$ such that for $\bm x$ in $\RR^2$ we have
\begin{align}
\bm \nu(\bm x) &= \bm\omega \times [\bm x,0] + \bm v \nonumber\\
&= [\,\underbrace{
 \bm{v}_T - \omega_N \bm {x}^\perp
 }_{\bm\nu_T(\bm x)}
 \, ,\, \underbrace{
 v_N + \langle \omega_T,\bm x^\perp \rangle
 }_{\nu_N(\bm x)}
\,],\label{eq:velo-decomp}
\end{align}
where for $\bm x=[a,b]$ in $\RR^2$, we denote by $\bm x^\perp = [b,-a]$ the vector rotated by an angle $\pi/2$ around the contact normal.
We denote $P^\perp$ the set of all $\bm x^\perp$ for $\bm x$ in $P$.

Dually, the contact force distribution \mbox{$\bm \rho(\bm x) = [\bm \rho_T(\bm x), \rho_N(\bm x)]$} for $\bm x$ in $P$, induces a wrench $\bm \lambda = [\bm m, \bm f] = [\bm{m}_T, m_N, \bm{f}_T, f_N]$ defined by the sum and the moment of the force distribution:
\begin{subequations}
\label{eq:wrench-decomp}
\begin{align}
 \bm f &= \int_P \bm \rho(\bm x) \, \mathrm{d}S_{\bm x}, \quad \bm m = \int_P [\bm x,0] \times \bm \rho(\bm x) \, \mathrm{d}S_{\bm x},\\
 f_N &= \int_P \rho_N(\bm x) \, \mathrm{d}S_{\bm x}, \quad \bm m_T = \int_P \rho_N(\bm x) \bm x^\perp \, \mathrm{d}S_{\bm x},\\
 \bm f_T &= \int_P \bm \rho_T(\bm x) \, \mathrm{d}S_{\bm x},\quad m_N = -\int_P \langle \bm \rho_T(\bm x), \bm x^\perp \rangle \, \mathrm{d}S_{\bm x} .
\end{align}
\end{subequations}

In some degenerate cases, $\bm \rho$ could represent a force distribution that has support over a sub-manifold $L \subset P$ or a discrete set of points.
Such cases arise, for example, when all the forces are concentrated on the boundary of the contact patch (e.g., a box in contact with a plane on one of its edges). 

\subsection{Signorini for point contact}
\label{subsec:point-signorini}
For each contact point identified by $\bm x$ in $P$, the nonpenetration imposes that the distance between the bodies must be positive.
Applying index reduction to such constraints as in \cite{acary2011} leads to $\nu_N(\bm x)\geq 0$.
The complementarity associated to this constraint is the so-called Signorini condition
\begin{align}
    0\leq \rho_N(\bm x) \perp \nu_N(\bm x) \geq 0. \label{eq:signo-point}
\end{align}
where $\rho_N(\bm x)$ is the dual variable of $\nu_N(\bm x)$ and corresponds to the normal force distribution.
The condition \eqref{eq:signo-point} is a cone complementarity problem with the cone $K = \RR_+$ that verifies $K^*=\RR_+$.

Physically, the Signorini condition encodes all the natural constraints. First, the two bodies cannot interpenetrate further, so $\nu_N\geq 0$. Second, the contact force can only be repulsive, so $\rho_N\geq 0$. Finally, if the bodies are separating $\nu_N > 0$, no force can happen (\mbox{$\rho_N = 0$}), the converse gives that if $\rho_N > 0$, then $\nu_N = 0$.
In other words, force and separation cannot occur simultaneously.

\subsection{Center of pressure}
\label{subsec:cop}
The center of pressure (CoP) is defined as the first moment of the normal force distribution:
\begin{align}
    \label{eq:center-of-pressure}
    \bm c_p=\frac{1}{\int_P \rho_N(\bm x) \, \mathrm{d}S_{\bm x}}\int_P \rho_N(\bm x) \bm x \, \mathrm{d}S_{\bm x}.
\end{align}
An implication of the point Signorini condition \eqref{eq:signo-point} on the whole contact patch $P$ is that $\rho_N(\bm x) \geq 0$ for every $\bm x$ in $P$. Thus, $\bm c_p$ is a convex combination of the elements of $P$, and we have that $\bm c_p$ belongs to the convex hull of $P$ denoted $\mathcal{C}(P)$.
This statement holds even when the distribution $\rho_N$ is supported on a lower-dimensional manifold $L \subset P$.

Another significant result about the center of pressure is that it relates to the tangential moment, i.e., the moment of the normal force distribution.
We denote by $\bm m_T(\bm c)$ the moment of the normal force distribution taken at a point of coordinate $\bm c$.
Using the so-called Varignon formula, we have
\begin{align}
 \bm m_T(\bm c) = \bm m_T - [\bm c,0]\times[\bm 0,f_N] =  \bm m_T - f_N \bm c^\perp.\label{eq:varignon-formulae}
\end{align}
The zero moment point $\bm c_\text{zmp}$ is the point where the normal force distribution exerts no moment: $\bm m_T(\bm c_\text{zmp}) = \bm 0$. Using \eqref{eq:varignon-formulae} and the wrench expression \eqref{eq:wrench-decomp}, the zero-moment point is well-defined when $f_N \neq 0$, and we have:
\begin{gather}
  \bm c_\text{zmp}^\perp= \frac{1}{f_N}\bm m_T = \frac{1}{\int_P \rho_N(\bm x) \, \mathrm{d}S_{\bm x}}\int_P \rho_N(\bm x) \bm x^\perp \, \mathrm{d}S_{\bm x}.
  \label{eq:positive-centerofpress}
\end{gather}
Thus, we have that $\bm c_p = \bm c_\text{zmp}$. This is a well-known result that has been extensively studied in robotics~\cite {Sardain01}.

%% file: tex/3_cone_complementarity.tex
\section{Conic Formulation, Duality and Signorini Law on a Contact Patch}
\label{sec:construction}

\begin{figure}[t!]
\vspace{.7em}
  \centerline{%
    \includegraphics[width=1.\linewidth, trim={0pt 0pt 60pt 0pt}, clip]{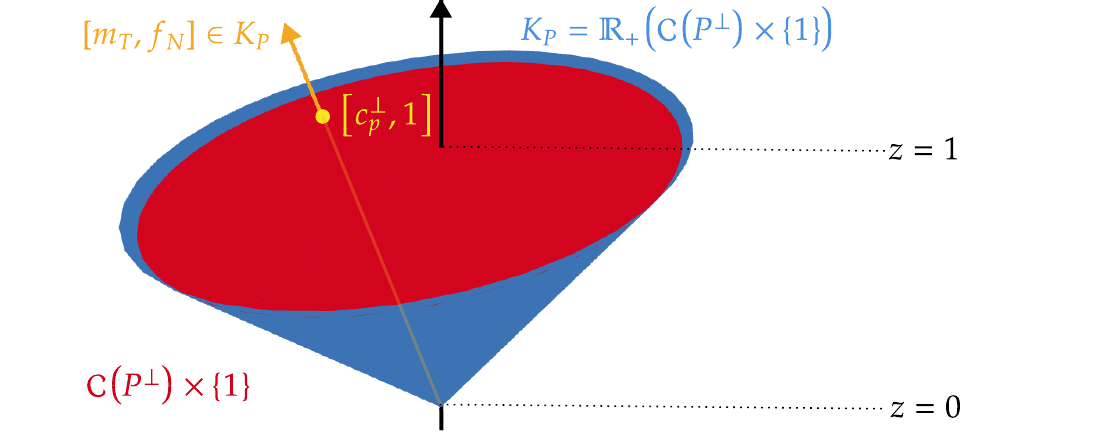}
 }
  \caption{\textbf{The cone $K_P$} is the set of all homogeneous coordinates of any point in $\mathcal{C}(P^\perp)$. In red we illustrate the $\mathcal{C}(P^\perp)\times \{1\}$ in $\RR^3$ in order to visualize well the cone $K_P$ in blue. An example of $[\bm m_T, f_N]\neq 0$ in $K_P$ is depicted in orange, which allows us to identify the center of pressure (up to a rotation) $\bm c_p$ in yellow. $\bm 0$ is also an element of $K_P$ with no associated center of pressure.
  }
  \label{fig:signocone-1}
\end{figure}

In this section, we derive the \textit{planar Signorini condition} which relates the contact wrench and the relative twist.
We demonstrate that the homogeneous coordinates of the center of pressure arises naturally and that a simple duality exists on the twist.
We also show that the planar Signorini condition is equivalent to the punctual Signorini condition for each point on the contact patch.
The section primarily focuses on the mathematical construction of the planar Signorini condition and its geometric interpretation; for completeness, a formal proof is provided in the Appendix.

\subsection{The CoP constraint as a cone constraint}
\label{subsec:cone-cop}
The CoP constraint described in Section~\ref{subsec:cop} states that if $f_N\neq 0$,  $\bm c_p$ must lie in the convex hull of the contact patch $P$, which we denote   $\mathcal{C}(P)$.
It comes from the repulsivity, which is the fact that $\rho_N(\bm x)\geq 0$ for all $\bm x$ in $P$.
Using the zero-moment point reformulation \eqref{eq:positive-centerofpress}, it is equivalent to $\frac{1}{f_N}\bm{m}_T$ belonging to $\mathcal{C}(P^\perp)$.

Another implication of the repulsivity is that if $f_N=0$, as $\rho_N(\bm x)\geq 0$, we necessarily have $\rho_N(\bm x)=0$ for almost any $x$, and by integration, since $\bm x$ is bounded, $\bm{m}_T = 0$.
Wrapping up, the repulsivity gives the two disjoint conditions on the contact wrench:
\begin{align}
\begin{cases} 
      \frac{1}{f_N}\bm{m}_T \in \mathcal{C}(P^\perp) &\text{if}\quad f_N\neq 0, \\
      \bm{m}_T = \bm 0 &\text{if}\quad f_N=0.
   \end{cases}\label{eq:case-constaint}
\end{align}

It follows that $[\bm{m}_T, f_N]$ is the homogeneous coordinates vector of the point $\bm c_p^\perp$ that belongs to $\mathcal{C}(P^\perp)$ as illustrated in Fig.~\ref{fig:signocone-1}.
Thus, the two conditions \eqref{eq:case-constaint} can actually be cast as a single condition in homogeneous coordinates.
This condition is that $[\bm{m}_T, f_N]$ belongs to the cone of homogeneous coordinates of any point in $\mathcal{C}(P^\perp)$, i.e., the convex cone of $\mathcal{C}(P^\perp)\times \{1\}$:
\begin{align}
    K_P = \RR_+\left(\mathcal{C}(P^\perp)\times \{1\}\right).
\end{align}
To be rigorous on the equivalence with the repulsivity, we prove the following Lemma.
\begin{mdframed}
\begin{lemma}[Repulsivity]\label{lem:repulsivity}
The point-wise repulsivity condition, i.e., $\rho_N(\bm x) \geq 0$ for every $\bm x$ in $P$, implies
\begin{align}
\begin{bmatrix}
\bm{m}_T , f_N
\end{bmatrix} \in K_P, \label{eq:patch-repulsivity}
\end{align}
and for every $[\bm{m}_T , f_N]$ verifying \eqref{eq:patch-repulsivity} there exist a compatible normal force distribution such that \mbox{$\rho_N(\bm x) \geq 0$} for every $\bm x$ in $P$.

\noindent The proof is given in the Appendix\,~\ref{proof:lemma1}.
\end{lemma}
\end{mdframed}

\subsection{Duality of Signorini and Twist Conic Constraint}
\label{subsec:signorini-duality}
\begin{figure}[t!]
\vspace{.7em}
  \centerline{%
    \includegraphics[width=1.\linewidth, trim={0pt 0pt 60pt 0pt}, clip]{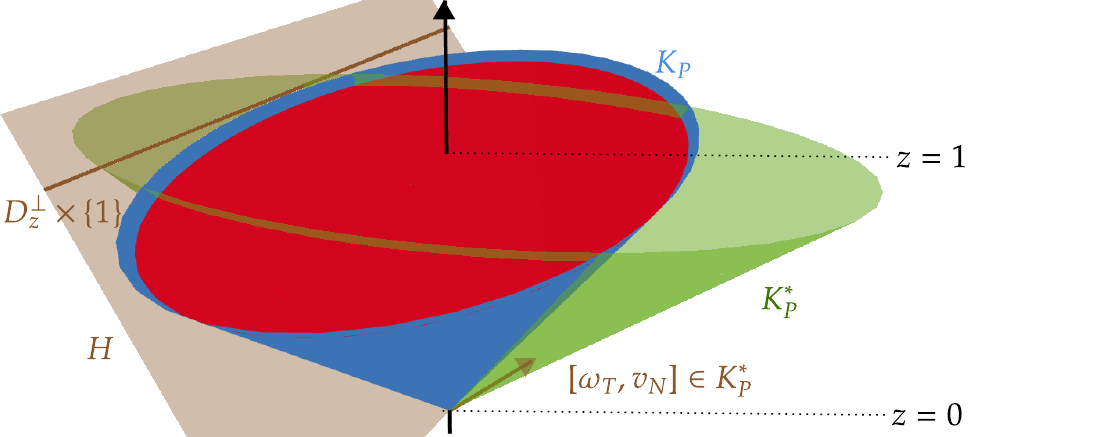}
 }
  \caption{\textbf{The dual $K_P^*$} is the set of all elements with a positive dot product with all the elements of $K_P$.
  It is the set of normals of oriented planes that leave $K_P$ on the positive side.
  This set is depicted in green.
  When $\bm{\omega}_T \neq \bm 0$, the plane intersects the contact plane ($z=1$) in a unique line, the zero normal velocity line (zero-line for short) $D_z$ (up to a $\pi/2$ rotation).
  In brown, we illustrate an element $[\bm{\omega}_T,v_N]$ in $K_P^*$ such that $\bm{\omega}_T \neq \bm 0$ with the associated zero-line.
  For any $v_N\geq 0$, $[\bm 0,v_N]$ is also an element of $K_P^*$ with no associated zero-line.
  }
  \label{fig:signocone-2}
\end{figure}
\begin{figure*}[t!]
\vspace{.7em}
  \centerline{%
    \includegraphics[width=.8\textwidth, trim={0pt 0pt 32pt 0pt}, clip]{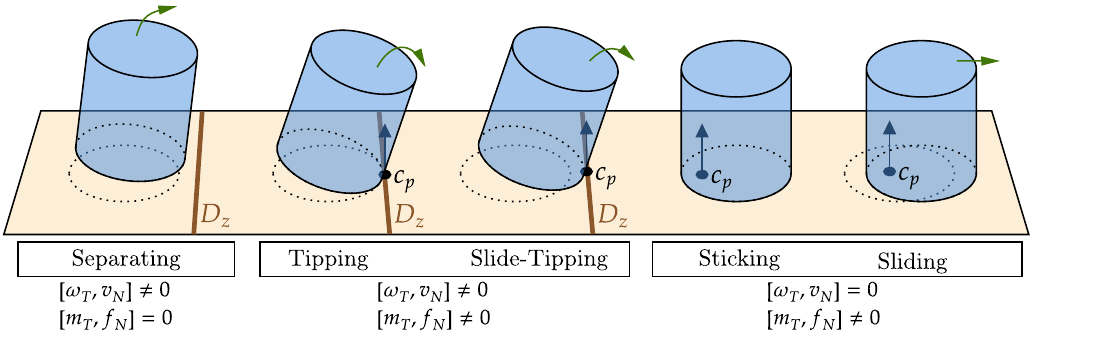}
 }
 \vspace{-.5em}
  \caption[The five different regimes in the contact patch problem.]{\textbf{The five different regimes in the contact patch problem.}
 If $\bm{\omega}_T$ is not zero, there is a line $D_z$ in the plane where $\nu_N(x) = 0$, and we represent it in brown.
 If $[\bm{m}_T,f_N]$ is not zero, the center of pressure $\bm c_p$ is well-defined by $\bm c_p^\perp = \bm{m}_T/f_N$  and we represent it as a black dot, together with an arrow representing $f_N$.
 We observe the three regimes of the cone complementarity of Planar Signorini, either $[\bm{\omega}_T, v_N]$ is zero: the contact is sticking or sliding depending on the friction, $[\bm{m}_T,f_N]$ is zero: the contact is breaking; or the center of pressure $\bm c_p$ must lie on the zero-line $D_z$, the contact is tipping or slide-tipping depending on the friction.
}
\vspace{-1.em}
  \label{fig:situation-objects}
\end{figure*}

The normal velocity field $\nu_N(\bm x)$ as defined in Eq.~\eqref{eq:velo-decomp} and recalled here:
\begin{align}
    \nu_N(\bm x) = v_N + \langle \bm{\omega}_T,\bm x^\perp\rangle,
    \label{eq:normal-velfield}
\end{align}
is an affine function of $\bm x$.
Using affine geometry, the nonpenetration constraint, i.e., $\nu_N(\bm x)\geq 0$ for every $\bm x \in P$, can be rewritten in homogeneous coordinates.
For any $[\bm y, \alpha]$ homogeneous coordinates of any $\bm x^\perp \in P^\perp$, we have that:
\begin{align}
     \langle [\bm{\omega}_T,v_N], [\bm y, \alpha]\rangle \geq 0.
\end{align}
In other words, $[\bm{\omega}_T,v_N]$ must have a positive dot product with any element of the cone of homogeneous coordinates of $P^\perp$.
It is precisely the definition of the dual cone of the cone generated by $P^\perp\times \{1\}$. 

Finally, the dual cone of a set is the dual cone of its convex hull, so we have that $[\bm{\omega}_T,v_N]$ belongs to $K_P^*$.
Formally, we have the following Lemma:
\begin{mdframed}
\begin{lemma}[Nonpenetration]\label{lem:non-pene}
The point-wise nonpenetration condition, i.e., $\nu_N(\bm x) \geq 0$ for every $\bm x$ in $P$, is equivalent to
\begin{align}
    \begin{bmatrix}
        \bm{\omega}_T, v_N
    \end{bmatrix} \in K_P^*. \label{eq:non-pene}
\end{align}

\noindent The proof is given in the Appendix\,~\ref{proof:lemma2}.
\end{lemma}
\end{mdframed}

The geometric interpretation of this result, depicted in Fig.~\ref{fig:signocone-2}, highlights a geometrically important object, the zero normal velocity line (zero-line for short) denoted by $D_z$ and characterized as:
\begin{subequations}
\label{eq:zero-line}
\begin{align}
    D_z &= \{\bm x \in \RR^2, \nu_N(\bm x) = 0\}\label{eq:zero-line-def}\\
    &= \{\bm x \in \RR^2, \langle \bm x, \bm{\omega}_T^\perp\rangle = v_N\}\label{eq:zero-line-carac}
\end{align}
\end{subequations}
Indeed, when $[\bm{\omega}_T,v_N]\neq \bm 0$, it can be seen as the normal to an oriented plane $H$ in $\RR^3$ and the nonpenetration condition becomes that $K_P$ must lie on the positive side of that plane (see Fig.~\ref{fig:signocone-2}).

\begin{figure}[b!]
\vspace{-.5em}
  \centerline{%
    \includegraphics[width=1.\linewidth, trim={0pt 0pt 60pt 0pt}, clip]{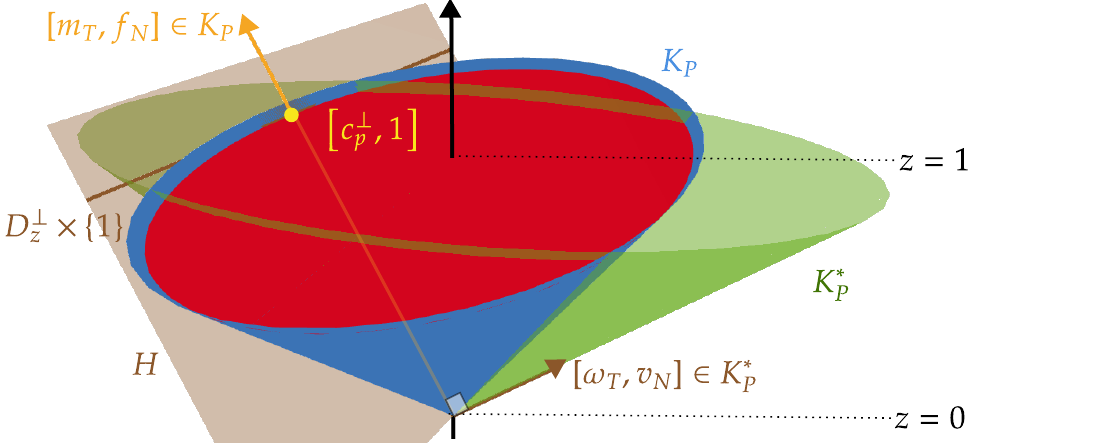}
 }
  \caption{\textbf{The nonzero twist and wrench case.} The complementarity in the planar Signorini condition \eqref{eq:patch-signorini} implies three regimes: \mbox{(i) $[\bm{\omega}_T,v_N]=\bm 0$} and $[\bm m_T,f_N]$ can be any element of $K_P$ as in Fig.~\ref{fig:signocone-1}, \mbox{(ii) $[\bm m_T,f_N]=\bm 0$} and $[\bm{\omega}_T,v_N]$ can be any element of $K_P^*$ as in Fig.~\ref{fig:signocone-2}, 
  or \mbox{(iii) $[\bm{\omega}_T,v_N] \neq \bm 0$} and $[\bm m_T,f_N] \neq \bm 0$ as depicted in this figure.
  In the last case, $[\bm{\omega}_T,v_N]$ (in orange) and $[\bm m_T,f_N]$ (in brown) lie on the boundary of their cone. We always have $f_N>0$ and $\bm{\omega}_T \neq 0$, so the CoP $\bm c_p$ (in yellow) and the zero-line $D_z$ (in brown) are well-defined and we have $\bm c_p \in D_z$.
  }
  \label{fig:signocone-3}
\end{figure}

When we analyze this back in the contact plane \mbox{($\RR^2 \times \{1\}$)}, we can distinguish between two cases: 
(i) $\bm{\omega}_T = \bm 0$ and the plane $H$ does not intersect the contact plane, the normal velocity field is the constant $v_N$, so the nonpenetration condition reduces to $v_N\geq 0$ as in the case of a single contact point. 
(ii) $\bm{\omega}_T \neq \bm 0$ and the plane $H$ intersects the contact plane in a unique oriented line $D_z^\perp$ with normal $\bm{\omega}_T$ depicted in Fig.~\ref{fig:signocone-2}. Rotating back by $\pi/2$, the nonpenetration condition becomes that the patch $P$ lies on the positive side of the oriented line $D_z$ with normal $-\bm{\omega}_T^\perp$, as illustrated in Fig.~\ref{fig:extended-center-of-pressure} and defined in \eqref{eq:zero-line}.
This characterization will serve to extend the notion of CoP.


\subsection{Planar Signorini: a Cone Complementarity Problem}
\label{subsec:patch-signorini}
Interestingly, the repulsivity and nonpenetration constraints for every point are exactly cone constraints on $[\bm{m}_T, f_N]$ and $[\bm{\omega}_T, v_N]$ that are dual to each other.
It remains to show that the complementarity holds.
This is not obvious, as when the punctual complementarity holds for each $\bm x$ in $P$, i.e., $\nu_N(\bm x)=0$ or $\rho_N(\bm x)=0$, we do not necessarily have $\nu_N$ or $\rho_N$ identically zero. In the trivial cases, we have either $[\bm m_T, f_N]=\bm 0$ or $[\bm{\omega}_T, v_N]=\bm 0$ and the complementarity holds immediately.
Yet, using some tools from convex analysis, we can show that in the non-trivial case, the CoP $\bm c_p$ and the zero-line $D_z$ are necessarily well-defined, and we must have $\bm c_p \in D_z$ as depicted in Fig.~\ref{fig:signocone-3}.
Because $[\bm m_T, f_N]$ are the homogeneous coordinates of the point $\bm c_p$ and $[\bm{\omega}_T,v_N]$ are the homogeneous coordinates of the oriented line $D_z$, $\bm c_p \in D_z$ is equivalent to:
\begin{align}
    \langle [\bm m_T, f_N], [\bm{\omega}_T,v_N]\rangle = 0.
\end{align}

Formally, we have the following proposition.
\begin{mdframed}
\begin{proposition}[Planar Signorini Condition]\label{prop:patch-signorini}
The punctual Signorini condition for every point of the patch, i.e., $0\leq \rho_N(\bm x) \perp \nu_N(\bm x)\geq 0$ for every $\bm x$ in $P$, implies that
\begin{align}
 K_P \ni \begin{bmatrix}
 \bm{m}_T,
 f_N
                \end{bmatrix} \perp
                \begin{bmatrix}
                \bm{\omega}_T,
 v_N
                \end{bmatrix} \in K_P^*, \label{eq:patch-signorini}
\end{align}
with $K_P = \RR_+ \left(\mathcal{C}(P^\perp)\times\{1\}\right)$.
Additionally, for every $[\bm{m}_T, f_N]$, $[\bm{\omega}_T, v_N]$ verifying \eqref{eq:patch-signorini} there exists a compatible normal force distribution $\rho_N$ such that \mbox{$0\leq \rho_N(\bm x) \perp \nu_N(\bm x)\geq 0$} for every $\bm x$ in $P$.\\[-0.5em]
\end{proposition}
\end{mdframed}
The proof, provided in the Appendix~\ref{proof:proposition1}, does not rely on any particular assumption regarding the force distribution.

It is interesting to note that the cone $K_P$ generalizes the cone $\RR_+$ of the contact-point case depicted in Eq.~\eqref{eq:signorini_condition_as_ccp}. 
Note also that if $P = \{\bm 0\}$, i.e. the patch only contains one point, $K = \{\bm 0\}\times\mathbb{R}_+$. This also means that $\bm{m}_T$ is zero, and \mbox{$K^* = \mathbb{R}^2 \times \mathbb{R}_+$}. In this case, $\bm{\omega}_T$ is not constrained, and we recover exactly the punctual Signorini condition.\\

\noindent\textbf{Physical interpretation.}
The nonpenetration, i.e., \mbox{$\nu_N(\bm x)\geq 0$} for all $\bm x$ in $P$, leads to two cases. Either $\bm{\omega}_T$ is zero and the normal velocity a nonnegative constant, or $[\bm{\omega}_T, v_N]$ is such that the zero-line $D_z$ does not intersect the interior of the contact surface and the patch $P$ lies on the positive half plane of $D_z$ where $\nu_N(\bm x)\geq 0$ as in Fig.~\ref{fig:extended-center-of-pressure}.
In the limit case, the zero-line intersects only the patch boundary, which is equivalent to $[\bm{\omega}_T, v_N]$ being on the boundary $\partial K_P^*$ (i.e., saturating the constraint).

On the other hand, for the contact wrench, the repulsivity, i.e., $\rho_N(\bm x)\geq 0$ for all $\bm x$ in $P$ leads to two cases. Either $[\bm{m}_T, f_N]$ is zero and the center of pressure is undefined, or $f_N\neq 0$ and the center of pressure is defined by $\bm c_p^\perp = \frac{1}{f_N}\bm{m}_T$ while belonging to $\mathcal{C}(P)$.
In the limit case, where the center of pressure is on the boundary of the surface, $[\bm{m}_T, f_N]$ is on the boundary $\partial K_P$.

The complementarity condition states that if the center of pressure is defined and is in the interior of the patch $P$, the normal velocity field must be uniformly zero. It is the case where $[\bm{m}_T, f_N] \neq 0$ and $[\bm{\omega}_T, v_N] = 0$, which physically corresponds to a sticking or sliding scenario depending on the tangential velocity field determined by  $[v_T, \omega_N]$ (see right part of Fig.~\ref{fig:situation-objects}).
On the other hand, if the normal velocity field is not uniformly zero and the patch $P$ strictly belongs to the positive half plane of $D_z$, the force distribution must be uniformly zero and $[\bm{m}_T, f_N]=0$. This is the case where $[\bm{m}_T, f_N] = 0$ and $[\bm{\omega}_T, v_N] \neq 0$, which physically corresponds to a separating scenario (see left part of Fig.~\ref{fig:situation-objects}).

The limit case where $[\bm{m}_T, f_N]$ and $[\bm{\omega}_T, v_N]$ are both non-zero is interesting. The complementarity condition states that the cone constraints are saturated. The center of pressure $\bm c_p$ is on the boundary of the patch $P$, and the zero-line of the normal velocity field intersects the boundary of the patch. More precisely, it imposes that the center of pressure must be in the intersection of the boundary of the patch and the zero-line.
Physically, it corresponds to a tipping scenario, as shown in the middle of Fig.~\ref{fig:situation-objects}, where the object is tipping around the zero-line $D_z$ and the normal force distribution degenerates to a submanifold of $D_z \cap P$, the intersection between the zero-line and the contact surface.
When the contact surface is strictly convex, as in Fig.~\ref{fig:situation-objects}, it always degenerates to a unique point.
In contrast, for a cube as in Fig.~\ref{fig:example-contact-patch}, the submanifold could be part of an edge of a cube.

\subsection{Extended Center of Pressure}\label{subsec:extended-cop}
\begin{figure}[t!]
\vspace{1em}
  \begin{center}
    \includegraphics[width=.7\linewidth, trim={0pt 0pt 170pt 0pt}, clip]{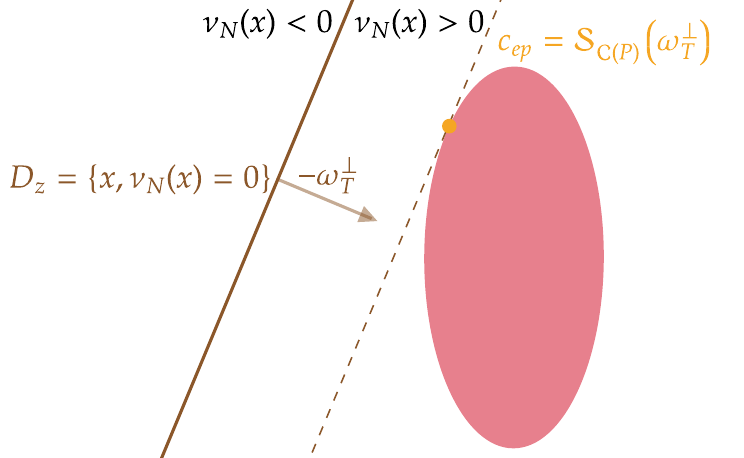}
    \caption{\textbf{Illustration of the extended center of pressure when $f_N=0$.} The zero-line $D_z$ in brown has normal $-\bm{\omega}_T^\perp$. The contact surface, indicated in red, is located in the positive half-plane. Here, the contact surface is strictly in the positive half-plane and $f_N=0$, so the CoP is not defined. However, we can define the extended CoP as the closest point, i.e., the most probable future CoP.
 }
    \label{fig:extended-center-of-pressure}
  \end{center}
\vspace{-1em}
\end{figure}
In the case where both $[\bm m_T,f_N]$ and $[\bm \omega_T,v_N]$ are nonzero, we saw that the CoP lies on $D_z \cap P$. Moreover, as described in Section~\ref{subsec:signorini-duality} and illustrated in Fig.~\ref{fig:extended-center-of-pressure}, $D_z$ is an oriented line with normal $-\bm{\omega}_T^\perp$ such that $P$ is in the positive half-plane.
Using the support function i.e., the set-valued function $\mathcal{S}$ that given a set $A$ of $\RR^2$ and a direction $d$ in $\RR^2$ is defined as $\mathcal{S}_A(d) = \operatorname{argmax}_{x\in A}\langle x,d\rangle$~\cite{Boyd2004, Rockafellar1970}, we have:
\begin{align}
    \bm c_p \in \mathcal{S}_{\mathcal{C}(P)}\left(\bm{\omega}_T^\perp\right),\label{eq:cp-belong}
\end{align}
meaning that $\bm c_p$ belongs to a subset of $\mathcal{C}(P)$ composed of the points closest to the zero-line $D_z$. When $P$ is strictly convex, and $\bm{\omega}_T \neq \bm 0$, $\mathcal{S}_{\mathcal{C}(P)}\left(\bm{\omega}_T^\perp\right)$ is a singleton. Though, interestingly, the condition \eqref{eq:cp-belong} holds even when $\bm{\omega}_T = \bm 0$, because $\mathcal{S}_{\mathcal{C}(P)}(\bm 0) = \mathcal{C}(P)$.

The condition \eqref{eq:cp-belong} suggests considering a set-valued extension of the center of pressure $\bm c_{ep}$, which is defined as:
\begin{align}
    \bm c_{ep}(\bm{m}_T,f_N,\bm{\omega}_T,v_N) = \begin{cases} 
      \left\{\frac{-1}{f_N}\bm{m}_T^\perp\right\} &\text{if}\quad f_N\neq 0 \\
      \mathcal{S}_{\mathcal{C}(P)}\left(\bm{\omega}_T^\perp\right) &\text{if}\quad f_N=0
   \end{cases}\label{eq:extended-cop}
\end{align}
Figure~\ref{fig:extended-center-of-pressure} depicts the extended center of pressure in the case of $f_N=0$.

A potential application of this concept could be to track the center of pressure immediately before and immediately after contact, to anticipate the torque required to maintain a stability criterion.
To detect the contact surface just before and just after the contact, it suffices to run the collision detection routine \cite{Gilbert1988,hppfcl} with a margin.

%% file: tex/5_conclusion.tex
\section{Conclusion}
\label{sec:conclusion}

In this paper, we introduced the \textit{planar Signorini condition}, a formulation that generalizes the classical point-contact Signorini condition to entire surfaces. This framework allows direct reasoning about the wrenches and twists compatible with a contact patch, without resorting to per-point discretization.

The proposed condition brings two main computational benefits over conventional point-based models:
(i)~it lowers the dimensionality of the associated constraint problem, and
(ii)~it enables physically sound treatment of surface contacts even for non-polygonal geometries.
Together, these properties allow for improved simulation of frictionless surface contact with improved efficiency.

Future work will incorporate this conic complementarity formulation into robotics simulators to more accurately and efficiently capture contact patches between surfaces, thereby reducing the computational burden of simulating complex physical interactions.  
Beyond simulation, the proposed conic complementarity model may prove especially valuable for locomotion and manipulation, where it naturally integrates with contact-implicit inverse dynamics methods \cite{menager2025contact}. 
Finally, an exciting next step is to extend the framework with a Coulomb friction model, further broadening its applicability to realistic robotic scenarios.

%% file: tex/proofs.tex
\subsection{Proof of Lemma 1}
\label{proof:lemma1}
Suppose $\rho_N(\bm x) \geq 0$ for every $\bm x$ in $P$.
By integration, we immediately have $f_N \geq 0$ and $f_N = 0$ if and only if $\rho_N(\bm x) = 0$ almost everywhere on $P$, which in turn implies by integration that $\bm{m}_T = 0$.
Now, suppose $f_N > 0$, using the zero moment point characterization of the center of pressure given in \eqref{eq:positive-centerofpress}, and the fact that with $\rho_N(\bm x) \geq 0$ for every $\bm x$ in $P$ the center of pressure is a convex combination of the elements of $P$, we have $\bm{c}_p$ belongs to $\mathcal{C}(P)$\footnote{This holds even when the distribution $\rho_N$ is supported on a lower-dimensional manifold $L \subset P$.} and in turn $\frac{1}{f_N}\bm m_T$ belongs to $\mathcal{C}(P^\perp)$.
Thus, there exists $y$ in $\mathcal{C}(P^\perp)$ such that $\bm{m}_T = f_N y$.

Adding the condition that $f_N \geq 0$ and that, when $f_N$ is zero, $\bm{m}_T$ is zero, we have the existence of $\alpha\geq 0$ such that
\begin{align}
    \begin{bmatrix}
 \bm{m}_T, f_N
    \end{bmatrix} = \alpha \begin{bmatrix}
 \bm y, 1
    \end{bmatrix},
\end{align}
which can be rewritten as $[\bm{m}_T, f_N] \in \mathbb{R}_+ (\mathcal{C}(P^\perp) \times \{1\}) = K_P$.\\

\noindent For the second statement of the Lemma:
\begin{itemize}
    \item if $f_N = 0$, the distribution $\rho_N(\bm x)=0$ is a valid distribution,
    \item if $ f_N \neq 0$, consider the center of pressure $\bm c_p$ defined by $\bm c_p^\perp = \frac{1}{f_N} \bm m_T$. As $[\bm{m}_T, f_N] \in K_P$, $\bm c_p$ is in $\mathcal{C}(P^\perp)$ and there exists $\bm x_1,\bm x_2$ in $P$ and $\alpha \in (0,1)$ such that $\bm c_p = \alpha \bm x_1 + (1-\alpha)\bm x_2$. The degenerated distribution that is supported only on $\{\bm x_1,\bm x_2\}$ and verifies $\rho_N(\bm x_1) = \alpha$,  $\rho_N(\bm x_2) = 1 - \alpha$ is a valid distribution.
\end{itemize}
\hfill$\blacksquare$

\subsection{Proof of Lemma 2}
\label{proof:lemma2}
Using the expression of the normal velocity field $\nu_N(\bm x)$, the point-wise nonpenetration condition gives that for every $\bm x$ in $P$:
\begin{align}
 0\leq \nu_N(\bm x) = \langle \bm{\omega}_T, \bm{x}^\perp \rangle + v_N = \left\langle \begin{bmatrix}
                \bm{\omega}_T,
 v_N
                \end{bmatrix},
                \begin{bmatrix}
 \bm{x}^\perp,
 1
                \end{bmatrix}
        \right\rangle.
\end{align}
The condition that $[\bm{\omega}_T,v_N]$ has a positive scalar for all $[\bm{x}^\perp,1]$ for all $\bm x$ in $P$ is exactly the condition \mbox{$[\bm{\omega}_T, v_N] \in K^*(P^\perp \times \{1\})$}.
Finally, we have:
\begin{align}
    K^*(P^\perp \times \{1\}) = K^*\left(\mathcal{C}(P^\perp) \times \{1\}\right) = K_P^*.
\end{align}
The converse is immediate.\hfill$\blacksquare$

\subsection{Proof of Proposition 1}
\label{proof:proposition1}
 Suppose $0\leq \rho_N(\bm x)\perp\nu_N(\bm x)\geq 0$ for every $\bm x$ in $P$, using Lemmas~\ref{lem:repulsivity} and \ref{lem:non-pene}, we have $[\bm{m}_T,f_N]\in K_P$ and $[\bm{\omega}_T,v_N]\in K^*_P$.
 Now let us prove that $\left\langle[\bm{m}_T,f_N],[\bm{\omega}_T,v_N]\right\rangle = 0$ by distinguishing three cases:
\begin{itemize}
    \item If $[\bm{m}_T,f_N]$ is zero, then we have $\left\langle[\bm{m}_T,f_N],[\bm{\omega}_T,v_N]\right\rangle = 0$ and the complementarity hold.
  
  \item If $[\bm{m}_T,f_N]$ is in the interior of $K_P$, then the center of pressure $\bm{c}_p$, defined by $\bm{c}_p^\perp = \frac{1}{f_N} \bm{m}_T$, is in the interior $\mathring{C}$ of $C = \mathcal{C}(P)$.
 Thus if the force distribution degenerates into a single point $\bm x$, we have $\bm x=\bm{c}_p \in \mathring{C}$ such that $\rho_N(\bm x) = f_N > 0$, and using the punctual Signorini condition $\nu_N(\bm x) = 0$.
 If the force distribution does not degenerate into a single point, there exist two points $\bm x_1$ and $\bm x_2$ in $P$ such that $\rho_N(\bm x_1) > 0$ and $\rho_N(\bm x_2) > 0$, with $\mathcal{C}\left(\{\bm x_1, \bm x_2\}\right) \cap \mathring{C} \neq \emptyset$.
 The punctual Signorini condition for $\bm x_1$ and $\bm x_2$ gives that $\nu_N(\bm x_1) = \nu_N(\bm x_2) = 0$, and for $0 \leq \alpha \leq 1$ such that $\bm x = \alpha \bm x_1 + (1-\alpha)\bm x_2 \in \mathring{C}$, with affine interpolation we have:
  \begin{align}
      \nu_N(\bm x) &= \nu_N(\alpha \bm x_1 + (1-\alpha) \bm x_2)\nonumber\\
      &= \alpha \nu_N(\bm x_1) + (1-\alpha) \nu_N(\bm x_2) = 0. \label{eq:conv-zero}
  \end{align}
 To summarize, if $[\bm{m}_T,f_N]$ is in the interior of $K_P$, there always exists $\bm x$ in $\mathring{C}$ such that $\nu_N(\bm x) = 0$.

 Because the normal velocity field $\nu_N:\RR^2\rightarrow \RR$ is affine and has value zero in a point located in the interior of a domain where $\nu_N$ is nonnegative, we have that $\nu_N$ is uniformly zero\footnote{
 Take a small radius $r > 0$ such that $B(\bm x, r) \subset C$, and take $\bm y$ in $B(\bm x, r)$. The point of coordinate $2\bm x - \bm y$ is in $B(\bm x, r)$ because $\|(2\bm x-\bm y) - \bm x\| = \|\bm x-\bm y\| \leq r$, and we have $\nu_N(2\bm x-\bm y) = \langle \bm{\omega}_T, (2\bm x-\bm y)^\perp \rangle + (2-1)v_N=2 \nu_N(\bm x) - \nu_N(\bm y)=- \nu_N(\bm y)$.
 We have $\nu_N(\bm y)\geq 0$ and $\nu_N(2\bm x - \bm y)=-\nu_N(\bm y)\geq 0$, so $\nu_N(\bm y)=0$.
 Thus, $\nu_N(\bm y) = 0$ for all $\bm y$ in $B(\bm x, r)$, and for any $\bm u$ in $B(\bm 0, r)$, $0 = \nu_N(\bm x+\bm u) - \nu_N(\bm x) = \langle \bm{\omega}_T, \bm u^\perp \rangle$, so $\bm{\omega}_T = \bm 0$ and finally taking $\bm u=\bm 0$, we have $0 = \nu_N(\bm 0) = v_N$.
 }. Thus we have $[\bm{\omega}_T, v_N] = 0$ and $\left\langle[\bm{m}_T,f_N],[\bm{\omega}_T,v_N]\right\rangle = 0$ and the complementarity holds.
 
  \item  If $[\bm{m}_T,f_N]$ is not zero and in the boundary of $K_P$, then the center of pressure $\bm{c}_p$, defined by $\bm{c}_p^\perp = \frac{1}{f_N} \bm{m}_T$, is in $\partial C$ the boundary of $C = \mathcal{C}(P)$.
 The center of pressure is a convex combination of elements of $P$, so the nonzero domain of the force distribution $\rho_N(\bm x)$ is included in the boundary $\partial{P}$ and is also included in the largest convex set $A$ such that $\bm{c}_p \in A \subset \partial C$ otherwise, the convex combination would be in the interior $\mathring{C}$.
 If $A$ is a singleton, $A = \{\bm{c}_p\}$, then the normal force distribution degenerates at one point, and we have $\rho_N(\bm{c}_p) = f_N > 0$, so using the Signorini condition on $\bm{c}_p$, we have $\nu_N(\bm{c}_p) = 0$.
 If $A = \mathcal{C}\left(\{\bm x_1, \bm x_2\}\right)$ is a segment, we have $\bm x_1$ and $\bm x_2$ in $P$ and the problem reduces to a 1-dimensional study over $A$.
 If $\bm{c}_p$ is on the boundary $\partial A = \{\bm x_1, \bm x_2\}$ of $A$, then the force distribution degenerates into one point, $\bm c_P$ is that point, and we have $\rho_N(\bm{c}_p) = f_N > 0$. 
 Finally, if $\bm{c}_p$ is in the interior $\mathring{A}$, either the force distribution degenerates to $\bm{c}_p$ and we have $\rho_N(\bm{c}_p) = f_N > 0$, or there exist $\bm y_1, \bm y_2$ in $A\cap \partial P$ such that $\rho_N(\bm y_1) > 0$, $\rho_N(\bm y_2) > 0$ and $\bm{c}_p$ is in $\mathcal{C}(\bm y_1, \bm y_2)$.
 The Signorini condition for $\bm y_1$ and $\bm y_2$ gives that $\nu_N(\bm y_1) = \nu_N(\bm y_2) = 0$ and using Eq.~\eqref{eq:conv-zero}, we have $\nu_N(\bm{c}_p) = 0$.
\end{itemize}

 To summarize, if $[\bm{m}_T,f_N]$ is not zero and in the boundary of $K_P$, we always have  $\nu_N(\bm{c}_p) = 0$.
 Using the expression for $\bm{c}_p^\perp$ in \eqref{eq:positive-centerofpress} and the expression $\nu_N$ in \eqref{eq:normal-velfield}, we have
  \begin{align}
      \nu_N(\bm{c}_p) = \langle \bm{\omega}_T, \bm{c}_p^\perp \rangle + v_N = \frac{1}{f_N} \langle \bm{\omega}_T, \bm{m}_T \rangle + v_N
  \end{align}
 Thus multiplying the previous equation by $f_N$, $\nu_N(\bm{c}_p) = 0$ gives
  \begin{align}
 0 &= \langle \bm{\omega}_T, \bm{m}_T \rangle + f_N v_N = \left\langle \begin{bmatrix}
 \bm{m}_T,
 f_N
              \end{bmatrix},
              \begin{bmatrix}
              \bm{\omega}_T,
 v_N
              \end{bmatrix}
      \right\rangle.
    \end{align}
 Summing up the three cases, if $0\leq \rho_N(\bm x)\perp\nu_N(\bm x)\geq 0$ for every $\bm x$ in $P$, we have $[\bm{m}_T,f_N]\in K_P$ and $[\bm{\omega}_T,v_N]\in K_P^*$ and always $\left\langle[\bm{m}_T,f_N],[\bm{\omega}_T,v_N]\right\rangle = 0$.
 Using the usual notations for CCP, we have:
  \begin{align}
 K_P \ni \begin{bmatrix}
 \bm{m}_T,
 f_N
                  \end{bmatrix} \perp
                  \begin{bmatrix}
                  \bm{\omega}_T,
 v_N
                  \end{bmatrix} \in K_P^*.\label{eq:ccp-proof}
  \end{align}

\noindent For the second statement of the Proposition, consider $[\bm{m}_T,f_N]$ and $[\bm{\omega}_T,v_N]$ verifying \eqref{eq:ccp-proof}, we have the following cases:
\begin{itemize}
    \item if $f_N = 0$, the distribution $\rho_N(\bm x)=0$ is valid,
    \item if $ f_N \neq 0$ and $[\bm \omega_T, v_N]=\bm 0$, any distribution valid for the $K_P$ constraint works as $\nu_N(\bm x)=0$ for any $\bm x$ in $P$.
    \item if $ f_N \neq 0$ and $[\bm \omega_T, v_N]\neq \bm 0$, consider the center of pressure $\bm c_p$ defined by $\bm c_p^\perp = \frac{1}{f_N} \bm m_T$. Due to the complementarity, we have $\bm c_p$ in $\partial \mathcal{C}(P)$. If $\bm c_p$ is in $P$, the degenerate distribution in $\bm{c}_p$ is compatible as using this distribution, $\rho_N(\bm x)=0$ for every $\bm x\neq \bm{c}_p$ in $P$, and $\left\langle[\bm{m}_T,f_N],[\bm{\omega}_T,v_N]\right\rangle = 0$ implies $\nu_N(\bm{c}_p)=0$. If $\bm c_p$ is not in $P$, then there exists $\bm x_1,\bm x_2$ in $\partial P$ and $\alpha \in (0,1)$ such that $\bm c_p = \alpha \bm x_1 + (1-\alpha)\bm x_2$. The line passing through $\bm x_1$ and $\bm x_2$ is the unique line containing $\bm c_p$ and not intersecting the interior of $\mathcal{C}(P)$, so it is necessarily $D_z$, and $\nu_N(\bm x_1) = \nu_N(\bm x_2) = 0$. The degenerated distribution that is supported only on $\{\bm x_1,\bm x_2\}$ and verifies $\rho_N(\bm x_1) = \alpha$,  $\rho_N(\bm x_2) = 1 - \alpha$ is a valid distribution.
\end{itemize}
\hfill$\blacksquare$